\documentclass[letterpaper, 10 pt, conference]{ieeeconf}  

\IEEEoverridecommandlockouts                              

\usepackage{cite}
\usepackage{graphics} 
\usepackage{epsfig} 
\usepackage{mathptmx} 
\usepackage{graphicx}
\usepackage[english]{babel}
\usepackage{blindtext}
\usepackage{times} 
\usepackage{amsmath} 
\usepackage[ruled,norelsize]{algorithm2e}
\usepackage{algpseudocode}
\usepackage{url}
\usepackage{amssymb}  
\usepackage{epstopdf}
\AppendGraphicsExtensions{.tiff}
\AppendGraphicsExtensions{.tif}
\makeatletter
\newcommand{\removelatexerror}{\let\@latex@error\@gobble}
\makeatother

\title{\LARGE \bf
Visual-Inertial Telepresence for Aerial Manipulation
}

\author{Jongseok~Lee$^{1}$, Ribin~Balachandran$^{1}$, Yuri~S.~Sarkisov$^{1, 2}$, Marco~De~Stefano$^{1}$, Andre~Coelho$^{1}$ \\ Kashmira~Shinde$^{1}$, Min~Jun~Kim$^{1}$, Rudolph~Triebel$^{1, 3}$ and Konstantin~Kondak$^{1}$
\thanks{$^{1}$ Institute of Robotics and Mechatronics, German Aerospace Center (DLR), Wessling, Germany. {\tt\small email:  jongseok.lee@dlr.de}}
\thanks{$^{2}$ Space CREI, Skolkovo Institute of Science and Technology (Skoltech), Moscow, Russia.}
\thanks{$^{3}$ Computer Vision Group, Technical University of Munich, Garching, Germany}}%

\begin{document}

\maketitle
\thispagestyle{empty}
\pagestyle{empty}


\begin{abstract}
This paper presents a novel telepresence system for enhancing aerial manipulation capabilities. It involves not only a haptic device, but also a virtual reality that provides a 3D visual feedback to a remotely-located teleoperator in real-time. We achieve this by utilizing onboard visual and inertial sensors, an object tracking algorithm and a pre-generated object database. As the virtual reality has to closely match the real remote scene, we propose an extension of a marker tracking algorithm with visual-inertial odometry. Both indoor and outdoor experiments show benefits of our proposed system in achieving advanced aerial manipulation tasks, namely grasping, placing, force exertion and peg-in-hole insertion.
\end{abstract}

\IEEEpeerreviewmaketitle

\section{Introduction}
Aerial manipulators exploit the manipulation capabilities of robotic arms located on a flying platform \cite{Ruggiero_survey}. These systems can be deployed for tasks that are unsafe and costly for humans. Some notable examples are repairing rotor blades of wind turbines and inspecting oil and gas pipelines in refineries. However, building an autonomous aerial manipulator \cite{Kondak2014, Ruggiero2015, Kim2018_compliance} poses several challenges to the current state-of-the-art robotic technologies. To this end, existing and close-to-market aerial manipulators are often tailored to a specific task such as contact inspection \cite{karen2019, Angel2019, Cuevas2019}.

An alternative is the remote control of an aerial manipulator (namely, aerial tele-manipulation). Aerial tele-manipulation, by having a human-in-the-loop, has an advantage that several demands on robot's cognitive modules can be replaced by its teleoperator. Furthermore, recent studies show promising results that indicate a possibility for deployment of such systems under an imperfect communication between the robot and the operator. For example, bilateral teleoperation with force feedback has been demonstrated in Kontur-2 mission \cite{spacejoystick} where a cosmonaut from the International Space Station successfully operated a robot on Earth. In aerial tele-manipulation, the works on force feedback \cite{Mohammadi2016} and shared control \cite{Franchi2012} can be notably found.

Additionally, 3D visual feedback is an another important aspect of aerial tele-manipulation systems for enhancing their manipulation capabilities. During our field experiments with such platforms, we experienced that a 2D visual feedback solely based on the live video streams is not sufficient to achieve precise manipulation tasks. Thus, we deduced that aerial telepresence systems must involve both real-time force and 3D visual feedback, which accurately displays the interactions of the robotic arm with the objects. Several studies confirm that a virtual environment where one can change its sight-of-view and provide haptic guidance (e.g. virtual fixtures) improves the system capabilities \cite{Falk2001InfluenceOT, Bettini2004, Huang2019}.

\begin{figure}
  \centering
  \includegraphics[width=0.485\textwidth]{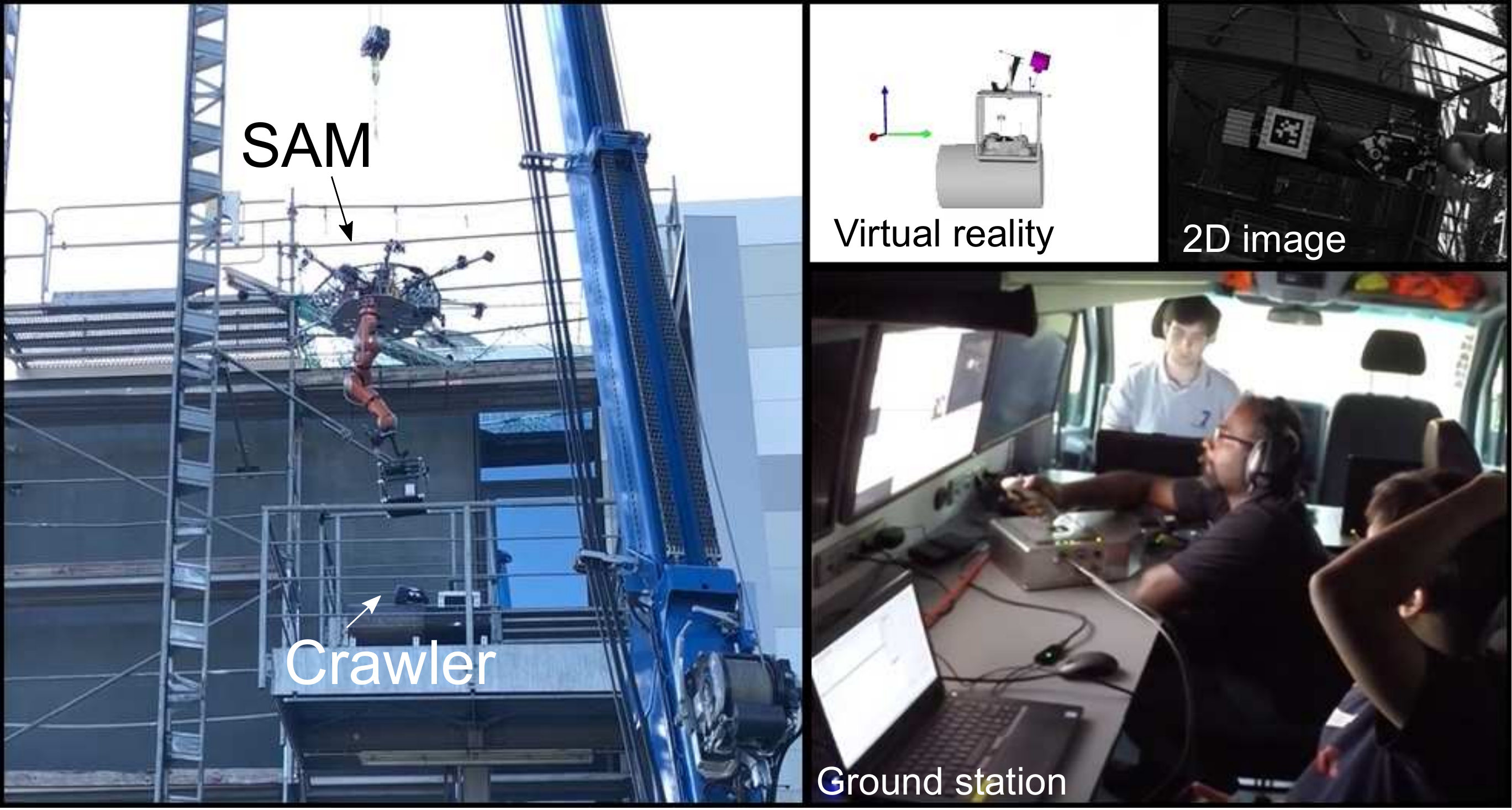}
 \caption{An illustration of the proposed concept. Our aerial robot SAM \cite{sam2019} is designed to achieve a manipulation task in a remote location where humans find it difficult to reach (see left side of the figure). Consequently teleoperator from a ground station does not have any visual contact to the scene. Therefore, the robot's onboard perception system must provide a visual feedback to the operator with both 2D and 3D information which overall enhance its manipulation capabilities (depicted in the right side).}
  \label{fig:sec1:1}
\end{figure}

Therefore, we propose an advanced visual-inertial telepresence system, which utilizes visual and inertial sensors to provide 3D visual feedback to the operator. The resulting system is equipped with a haptic feedback and a virtual reality with virtual fixtures. In particular, for creating the 3D display of a remote scene, we consider an object localization approach where an object database and a marker tracking algorithm are used. As existing marker tracking methods did not suffice our requirements in terms of robustness and run-time, we propose a new object tracking algorithm by extending ARToolKitPlus \cite{artoolkit} with onboard visual-inertial odometry (VIO). Lastly, an extension of the framework to multiple objects is also addressed for pick-and-place tasks.

The proposed concept is tightly integrated to a collision-safe aerial manipulator called cable-Suspended Aerial Manipulator (SAM\cite{sam2019}). In particular, the main scenario of interest is to deploy and retrieve an inspection robotic crawler (as illustrated in Fig.  \ref{fig:sec1:1}). This scenario, which was designed under the scope of EU project AEROARMS, is relevant to inspection and maintenance of gas and oil pipelines in refineries\cite{aeroarms}. It involves grasping, placing and pressing tasks which need to be performed by a remotely located operator. The proposed algorithm is validated indoors and a peg-in-hole task with a margin of error less than 1cm is studied, which further displays SAM's advanced manipulation skills.

In summary, our main contributions are as follows.
\begin{itemize}
    \item A visual-inertial telepresence system for aerial manipulation where a new object localization approach is proposed for creating virtual reality of the remote scene.
    \item An extension of ARToolKitPlus \cite{artoolkit} with onboard VIO for improving its run-time and robustness.
    \item Experimental validations showing advanced manipulation skills with SAM for the first time. In particular, our field experiments indicate overall system as a viable option for inspection and maintenance applications.
\end{itemize}
Experiments can be seen in the video: \url{https://www.youtube.com/watch?v=onOc05Ymxzs}.

\subsection{Related Works} Several researchers aimed to provide 3D information of the remote scene for tele-manipulation. For this, 3D reconstruction techniques have been notably applied so far \cite{Gerd1999, ni_song_xu_li_zhu_zeng_2017, Dejing2019, Leeper2012icra, Ryden2013} where they aimed to create 3D visualization of an unknown environment. However, their applicability to our use-case is limited as the scene has to be mapped first, and then pre-processed for coping with the noisy 3D vision data. Unlike these methods, our approach differs as we use object localization algorithms. Two benefits are: (i) a real-time display is possible, and (ii) the framework can also be extended to a pick-and-place task, which requires the visualization of both the hand-held object and the target of placement. The later is difficult with the existing methods when the hand-held object is not rigidly fixed to a gripper. A recent work AeroVR \cite{yashin2019aerovr} uses a similar concept to ours. While the system demonstrates an inspiring way to also include tactile feedback, the scope differs as AeroVR uses VICON system for indoor usage.

For object localization, learning-based \cite{learning_1, learning_2, Sundermeyer} and geometry-based \cite{Wang2016, artoolkit} approaches can be found. Recent learning-based methods with deep neural networks can be broadly formulated with either explicit \cite{learning_1} or implicit \cite{Sundermeyer} representations. However, we do not consider machine learning approaches as the assumption that the test data distribution to come from training distribution is routinely violated in the context of field robotics. Within the geometric methods, Fidicual marker systems (based on creating artificial features on the scene) are widely used in robotics for ground truths \cite{Wang2016}, applications where environments are known \cite{Malyuta2019}, simplifying the perception problem in lieu of sophistication \cite{Laiacker2016} and calibration \cite{nissler18simultaneous}. However, as we aim for creating the real-time virtual reality, our use-case provides stringent requirements on their limitations in run-time and inherent time-delays. Note that authors \cite{predictive19} show that coping with time delays in the display improves the performance of the tele-operation. Furthermore, as we use hand-eye cameras, our localization method should be robustness to loss-of-sight as the camera is not guaranteed to see the markers during the operations. Robustness is important when using haptic guidance or virtual fixtures for example, where inaccurate haptic feedback can cause negative effects in terms of the manipulation performance \cite{7047839, Boessenkool}.
\section{Cable Suspended Aerial Manipulator}
\label{sec:sam}
\begin{figure}
  \centering
  \includegraphics[width=0.45\textwidth]{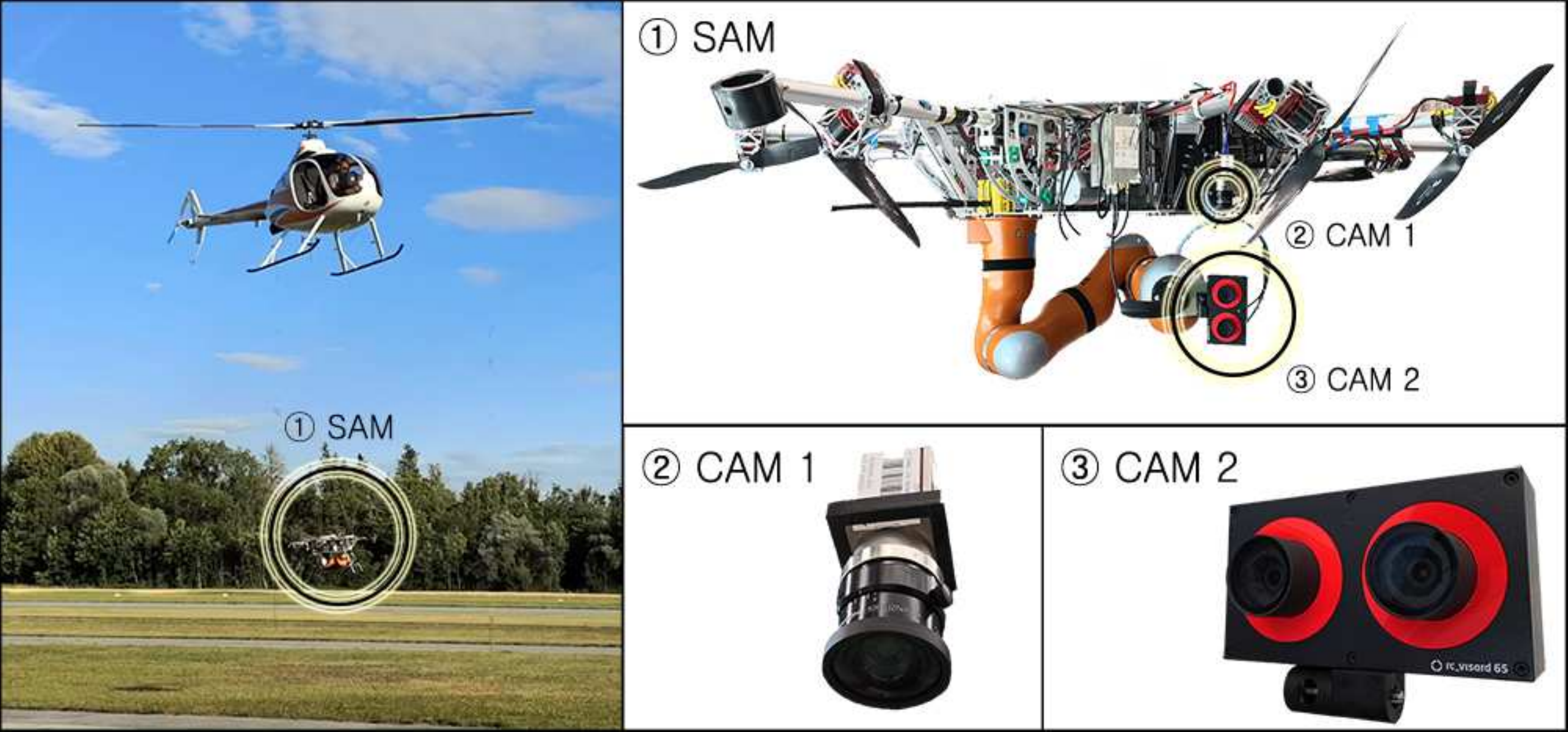}
 \caption{Illustration of our collision-safe aerial manipulation concept; SAM with helicopter as an aerial carrier (left). Both hand-eye and eye-to-hand cameras are now integrated (right). We denote CAM1 as mako and CAM2 as hand-eye camera (hc for brevity).}
  \label{fig:sec2:1}
\end{figure}
\subsubsection{Robot hardware}
An aerial manipulator SAM \cite{sam2019} is a complex flying robot composed of an aerial carrier, a cable-Suspended platform and a 7 degrees of freedom (DoF) industrial robotic arm KUKA LWR \cite{kuka}. An aerial carrier (e.g. crane, manned/unmanned helicopter\footnote{The purpose of the aerial carrier is to transport the system and hover. We use a crane in this study which also provides better safety, versatility, robustness and applicability for our considered application scenario.}) provides means to transport the robotic platform to a location (see Fig. \ref{fig:sec2:1}). Then, a platform suspended to the carrier performs balancing act by autonomously damping out the disturbances induced by the carrier and the manipulator. This oscillation damping control is performed using eight omni-directional propellers and three winches as its actuators. Design and control aspects of SAM have been presented previously in \cite{sam2019}.

\subsubsection{Sensors choices and integration}
Relevant sensors for realizing our vision-based telepresence system are as follows. KUKA LWR \cite{kuka} is equipped with torque and position sensors as its \textit{proprioceptive} sensors. Each joint contains a torque sensor, incremental and absolute position sensors which measure its joint torques and angles. Furthermore, SAM is equipped with optical devices as its \textit{exteroceptive} sensors. As shown in Fig. \ref{fig:sec2:1}, a monocular camera (Allied-vision Mako) is installed on the frame of the platform to display the overall operational space of the robotic arm. This is because the operator prefers eye-to-hand view which is more natural to a human. The camera provides high resolution images of 1292 by 964 px at 30Hz. Additionally, a stereo camera is integrated near the tool-center-point (tcp). Accuracy of the ficidual marker systems depends on the distance and its size which justifies the integration of a hand-eye camera \cite{Wang2016}. We use a commercial 3D vision sensor that provides built-in VIO. Rcvisard provides 1280 by 960 px images at 25Hz and VIO estimates can be acquired at 200Hz. Details on VIO algorithm can be found in \cite{rcvisard}

\subsubsection{Haptic device}
A portable and space-qualified haptic device, the Space Joystick \cite{spacejoystick} has been integrated to teleoperate the LWR located on SAM remotely.

\section{Vision-Inertial Aerial Telepresence}
\label{sec:core}
\subsection{3D Visual Feedback with Object Localization}
\begin{figure}
  \centering
  \includegraphics[width=0.5\textwidth]{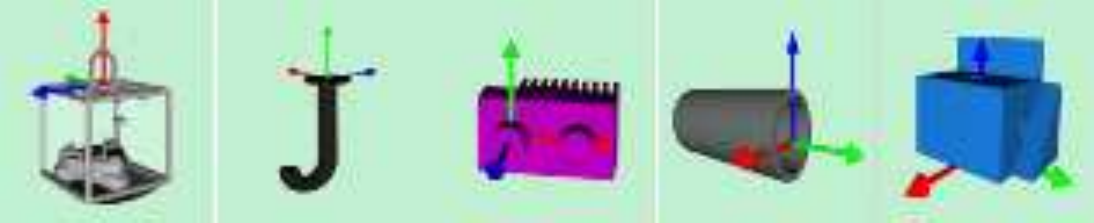}
 \caption{An example of pre-generated object database.}
  \label{fig:sec3:2}
\end{figure}

The aim is virtually displaying the robot and the objects so that an operator can tele-manipulate remotely. If done in real-time, the operator can \textit{see} the virtual remote scene and perform the tasks. Here, accuracy is crucial as the virtual world has to closely match the real remote scene. In our approach, we realize such 3D visual feedback using cameras, object localization algorithms and known object database (see Fig.  \ref{fig:sec3:2}). Once objects to be actively manipulated are known a-priori, the essence of the problem simplifies to computing relative transformation of an objects with respect to the camera $\textbf{T}_{\text{object}}^{\text{hc}}(t)$ and robot's tcp $\textbf{T}_{\text{object}}^{\text{tcp}}(t)$. Here, t denotes time. A fixed transformation $\textbf{T}_{\text{hc}}^{\text{tcp}}$ can be precisely estimated from CAD models or hand-eye calibration \cite{Strobl2006}.
\begin{equation}
\label{sec:3:2:eq1}
    \textbf{T}_{\text{object}}^{\text{tcp}}(t) = \textbf{T}_{\text{hc}}^{\text{tcp}} \textbf{T}_{\text{object}}^{\text{hc}}(t)
\end{equation}

In this way, one can exploit object localization methods based on fiducial markers systems. These systems are widely adopted in robotics community and have been used as ground truths for its accuracy \cite{Wang2016}. While learning-based pose estimation methods \cite{Sundermeyer} can be leveraged under the same framework (for several applications where markers are not readily available), we limit our scope to validating the virtual reality concept in lieu of sophisticated object localization methods. Note that we use Instant Player \cite{hulin2012} for creating the display as it supports various hierarchies of a scene graph. Using a nested hierarchy, relative transformation between an object and tools can be routed to display the scene, while a flat hierarchy can be used to extend the framework in order to display multiple objects and tools.

\begin{figure}
  \centering
  \includegraphics[width=0.45\textwidth]{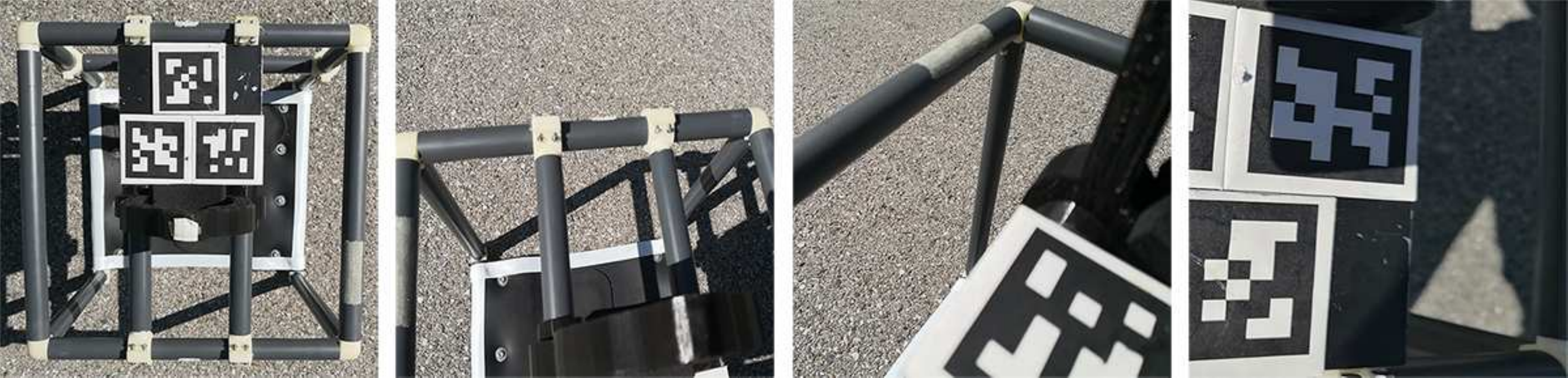}
 \caption{Failure modes of fidicual marker system in the field experiments. The figure shows a nominal case (left), and failure modes namely lost of sight and shadow occlusion (others).}
  \label{fig:sec3:3}
\end{figure}

However, fiducial markers systems and their extensions \cite{artoolkit, Wang2016, Malyuta2019, Laiacker2016} have also significant drawbacks. It arises as we consider floating base manipulation outdoors. For example, shadows are inevitable for outdoor experiments and once it destroys certain shapes of the tags, the methods would naturally fail as its assumptions on the artificial visual features are violated. Similarly, the hand-eye camera (hc) can lose the view on the marker as the manipulator and the base can move rapidly. These failure modes (reported in Fig.  \ref{fig:sec3:3}) have consequences on the mission success rates. This is because it is difficult for the operator to remotely perform precise manipulation with live streams of 2D images. Eye-to-Hand views typically suffer from the occlusions of the grasping points by the robotic arm (also found in humanoid robots) and lacks depth information. Lastly, time delays that are inherent in these systems must be corrected in order to create a real-time virtual display of the scene.

For tackling these problems we propose Algorithm 1 for which multiple tags are placed on an object with a target tag x. The algorithm initializes by detecting all the tags (we denote multiART+ which is based on ArtoolKitPlus \cite{artoolkit}), and saving their relative poses to the target (tag$\_$init). While the process is running, k detected tags and their IDs are counted (counter$\_$multiART+). If all the tags are detected, n+1 pose estimates of the target tag x can be computed by transforming pose estimates of non-target tags $T_{\text{y}}^{\text{hc}}$ and their relative transformation to the target tag $T_{\text{x}}^{\text{y}}$ (trafo3d). Then, RANSAC \cite{ransac} is applied to these estimates to remove outliers, and then averaging to reduce variance (ransac$\_$avg). Then, relative transformations are updated by applying RANSAC for the saved estimates, and averaging. In case atleast one tag is detected, the same step is applied to estimate the target tag x. These steps have advantages that (1) accuracy and orientation ambiguity of ArtoolKitPlus can be improved with RANSAC, and (2) the algorithm is robust to loss-of-sight of a target (similar to \cite{Laiacker2016, Nissler, Malyuta2019}). 

\begin{figure}
  \centering
  \includegraphics[width=0.5\textwidth]{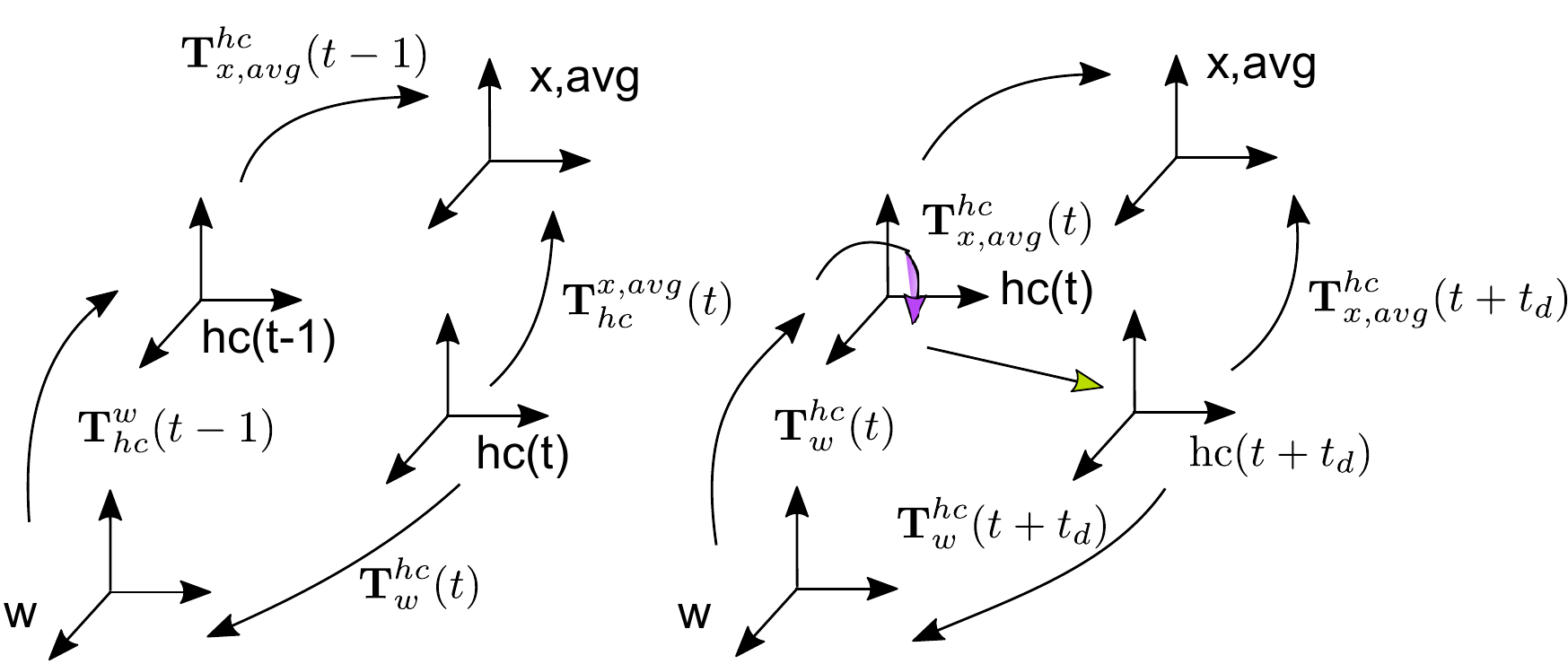}
 \caption{Illustration of \eqref{eq:vio_integrate} and \eqref{eq:vio_compensator}. Left: \eqref{eq:vio_integrate} uses VIO position and orientation estimates of camera motion to still estimate the object (denoted x,avg) when marker is not detected. Right: \eqref{eq:vio_compensator} uses linear (yellow arrow) and angular velocity (blue arrow), and computed time delay $t_d$ to predict the motion of the camera in $t+t_d$ seconds.}
  \label{illustration}
\end{figure}

However, the algorithm must be robust to loss-of-sight on all the tags, as we consider object tracking for floating base manipulators. Algorithm 1 addresses this problem by integrating VIO estimates of camera motion with respect to its inertial coordinate $\textbf{T}_{\text{hc}}^{\text{w}}(t)$. If no tags are detected, (\ref{eq:vio_integrate}) can be used to still estimate the target $\textbf{T}_{\text{x,avg}}^{\text{hc}}(t)$ (vio$\_$integrate). In (\ref{eq:vio_integrate}), $\textbf{T}_{\text{w}}^{\text{hc}}(t)\textbf{T}_{\text{hc}}^{\text{w}}(t-1)$ is a relative transformation of camera motion from time t-1 to t and assumes a static object.
\begin{equation}
\label{eq:vio_integrate}
\textbf{T}_{\text{x,avg}}^{\text{hc}}(t)  = \textbf{T}_{\text{w}}^{\text{hc}}(t)\textbf{T}_{\text{hc}}^{\text{w}}(t-1)\textbf{T}_{\text{x,avg}}^{\text{hc}}(t-1)   
\end{equation}
In a similar fashion, the delay of the system $t_d$ can be computed (delay$\_$computation) and corrected with VIO algorithm by using (\ref{eq:vio_compensator}) (vio$\_$delay$\_$compensator). The herein delay is present in any perception system (e.g. rectifying an image) and fiducial marker systems (they are not real-time). In (\ref{eq:vio_compensator}), $\textbf{T}_{\text{hc}}^{\text{w}}(t)$ and $\textbf{T}_{\text{x,avg}}^{\text{hc}}(t)$ are computed using VIO and multi-tag tracking. On the other hand, $\textbf{T}_{\text{w}}^{\text{hc}}(t+t_d)$ can be computed using linear and angular velocity estimates of VIO, multiplied by the delay time $t_d$.
\begin{equation}
\label{eq:vio_compensator}
\textbf{T}_{\text{x,avg}}^{\text{hc}}(t+t_d)  = \textbf{T}_{\text{w}}^{\text{hc}}(t+t_d)\textbf{T}_{\text{hc}}^{\text{w}}(t)\textbf{T}_{\text{x,avg}}^{\text{hc}}(t)   
\end{equation}
These two steps have several advantages. The algorithm is robust to failure modes of fidicual marker systems (see Fig.  \ref{fig:sec3:3}) as it copes with missing tag detection, and time delays are incorporated by using velocity signals and computed delay time. Furthermore, maximum run-time of the algorithm can be pushed to that of VIO data. The algorithm deals also with drifts of VIO estimates by using relative motion estimates only when the tag detection is lost. Note that the method is one way to use commodity vision sensors with VIO modules in order to further improve performance. Illustration of these two steps are found in Fig. \ref{illustration}.
\begin{figure}
 \removelatexerror
  \begin{algorithm}[H]
   \caption{Robust marker localization}
   \textbf{Input}: 
   Image $\textit{I}$, target marker ID x, n multi marker IDs y and mapping to object $\textbf{T}_{\text{object}}^{\text{x}}$. \\
   \textbf{Output}: 
   Pose of the object $\textbf{T}_{\text{object}}^{\text{stereo}}$(t). \\
   \textbf{Algorithm}: \\
   $\textbf{T}_{\text{x}}^{\text{hc}}$(0), $\textbf{T}_{\text{y}_1}^{\text{hc}}$(0), ..., $\textbf{T}_{\text{y}_n}^{\text{hc}}$(0) $\leftarrow$ multiART+($\textit{I}$); \\
   $\textbf{T}_{\text{x}}^{\text{y}_1}$, $\textbf{T}_{\text{x}}^{\text{y}_2}$, ..., $\textbf{T}_{\text{x}}^{\text{y}_n}$ $\leftarrow$ tag$\_$init($\textbf{T}_{\text{x}}^{\text{hc}}$(0),$\textbf{T}_{\text{y}_1}^{\text{hc}}$(0), ..., $\textbf{T}_{\text{y}_n}^{\text{hc}}$(0))\\
   \While(){\emph{object$\_$localization == True}}
   {
      k, id $\leftarrow$ counter$\_$multiART+($\textit{I}$); \\
      \uIf{\emph{k == n+1}}{
      $\textbf{T}_{\text{x}}^{\text{hc}}$(t), ..., $\textbf{T}_{\text{x,yn}}^{\text{hc}}$(t) $\leftarrow$ trafo3d($\textbf{T}_{\text{x}}^{\text{hc}}$(t), $\textbf{T}_{\text{y}}^{\text{hc}}$(t), $\textbf{T}_{\text{x}}^{\text{y}}$)\;
      $\textbf{T}_{\text{x,avg}}^{\text{hc}}$(t) $\leftarrow$ ransac$\_$avg($\textbf{T}_{\text{x}}^{\text{hc}}$(t), ..., $\textbf{T}_{\text{x,yn}}^{\text{hc}}$(t))\;
      $\textbf{T}_{\text{x}}^{\text{y}_1}$, $\textbf{T}_{\text{x}}^{\text{y}_2}$, ..., $\textbf{T}_{\text{x}}^{\text{y}_n}$ $\leftarrow$ tag$\_$init$\_$update($\textbf{T}_{\text{x,pre}}^{\text{y}}$, $\textbf{T}_{\text{x}}^{\text{y}}$)\;
      }
      \uElseIf{\emph{0 $<$ k $<$ n+1}}{
      \uIf{\emph{x $\in $ id == False}}{
      $\textbf{T}_{\text{x,y1}}^{\text{hc}}$(t), ..., $\textbf{T}_{\text{x,yn}}^{\text{hc}}$(t) $\leftarrow$ trafo3d($\textbf{T}_{\text{y}}^{\text{hc}}$(t), $\textbf{T}_{\text{x}}^{\text{y}}$)\;
      $\textbf{T}_{\text{x,avg}}^{\text{hc}}$(t) $\leftarrow$ ransac$\_$avg($\textbf{T}_{\text{x,y1}}^{\text{hc}}$(t), ..., $\textbf{T}_{\text{x,yn}}^{\text{hc}}$(t))\;
      }
      \Else{
      $\textbf{T}_{\text{x}}^{\text{hc}}$(t), ..., $\textbf{T}_{\text{x,yn}}^{\text{hc}}$(t) $\leftarrow$ trafo3d($\textbf{T}_{\text{y}}^{\text{hc}}(t)$, $\textbf{T}_{\text{x}}^{\text{y}}$)\;
      $\textbf{T}_{\text{x,avg}}^{\text{hc}}$(t) $\leftarrow$ ransac$\_$avg($\textbf{T}_{\text{x}}^{\text{hc}}$(t), ..., $\textbf{T}_{\text{x,yn}}^{\text{hc}}$(t))\;
      }
      }
      \Else{
      $\textbf{T}_{\text{x,avg}}^{\text{hc}}$(t) $\leftarrow$ Eq. (2)\;
      }
      $t_d$ $\leftarrow$ delay$\_$computation() \\
      $\textbf{T}_{\text{x,avg}}^{\text{hc}}(t+t_d)$ $\leftarrow$ Eq. (3) \\
      $\textbf{T}_{\text{object}}^{\text{hc}}(t)=\textbf{T}_{\text{x,avg}}^{\text{hc}}(t+t_d)\textbf{T}_{\text{object}}^{\text{x}}$
   }
  \end{algorithm}
\end{figure}

\subsection{Extension of 3D Visualization to Multiple Objects}
For tasks such as placing, virtually displaying multiple objects and their relative pose is required. For example, if an operator would like to place a cage (with inspection robot inside) on a pipe which have roughly the same dimension, the virtual reality should reflect it by displaying the pipe, the cage, and the orientation changes of the cage with respect to TCP (e.g. a hook). With 3D reconstruction methods, this is difficult as one explores the environment for mapping and process the noisy data points for displaying. In our system, we tackle this challenge by using the hand-eye camera to estimate the orientation of the held object, while the eye-to-hand camera estimates the pose of other objects (e.g. a pipe). Then, the forward kinematics are leveraged as given below.
\begin{equation}
\label{sec:3:3:eq1}
\textbf{T}_{\text{object,2}}^{\text{object,1}}(t) = \textbf{T}_{\text{mako}}^{\text{object,1}}(t) \textbf{T}_{\text{base}}^{\text{mako}}\textbf{T}_{\text{tcp}}^{\text{base}}(t)\textbf{T}_{\text{hc}}^{\text{tcp}}\textbf{T}_{\text{object,2}}^{\text{hc}}(t)
\end{equation}
In (\ref{sec:3:3:eq1}), transformation from the base to eye-to-hand camera (mako) $\textbf{T}_{\text{base}}^{\text{mako}}$ and tcp to hand-eye camera $\textbf{T}_{\text{hc}}^{\text{tcp}}$ can be computed using hand-eye calibration \cite{Strobl2006}. $\textbf{T}_{\text{mako}}^{\text{object,1}}$ is essentially updating the local base frame, and the forward kinematics of the robotic arm $\textbf{T}_{\text{tcp}}^{\text{base}}$ is typically accurate. $\textbf{T}_{\text{object,2}}^{\text{hc}}$ displays the pose of the held object. For this, one can use only multi-marker tracking without linear velocity integration. This is because markers can always made visible when the objects are held by the robot.

\subsection{Force Feedback with Space Joystick and LWR}
\begin{figure}
  \centering
  \includegraphics[width=0.5\textwidth]{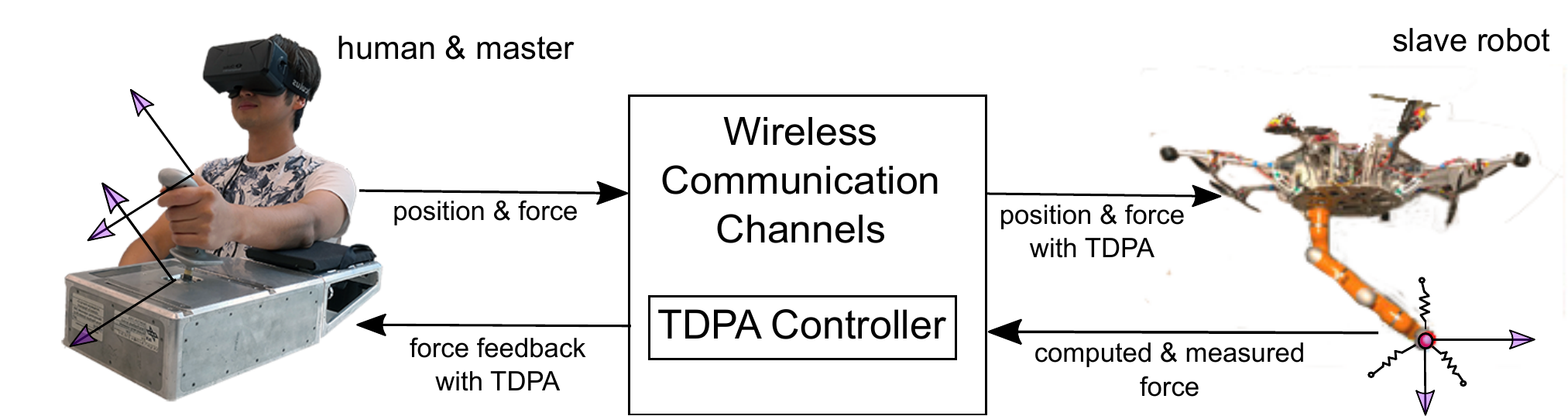}
 \caption{Controller overview. Communication time delays, packet loss and jitter can cause instability of the overall system. For coping with this, TDPA is used for force feedback tele-manipulation.}
  \label{fig:sec3:1}
\end{figure}

The controller design must ensure a stable bilateral tele-manipulation with force feedback. The main technical challenge is to deal with communication time delays, packet loss and jitters, which can cause instability of the system. For tackling this, a four channel architecture with time-domain passivity approach (proposed in \cite{spacejoystick}) has been used. A schematic of the system is shown in Fig. 5 and it is briefly explained as follows. The human operator sends both position (velocity analogously) and force signals from the master device (Space Joystick) to the slave (KUKA LWR mounted on the SAM). As these signals pass through communication channels (in the considered scenario, a wireless communication), they will get affected by time delay. To ensure stable tele-manipulation, we employ time domain passivity approach (TDPA \cite{Hannaford2001TimeDP}). Readers can refer to \cite{spacejoystick} for more details and implementations. 

\subsection{Haptic Guidance with Virtual Fixtures}
On top of real-time virtual reality and haptic device, another aspect of our telepresence system is haptic guidance via virtual fixtures \cite{Bettini2004}. In this work, the virtual fixtures are implemented as artificial walls that guide the motion of the slave to the desired target point. If the teleoperator tries to move the slave device outside these walls, artificial forces are activated to limit the motion of tcp (slave) and also to provide haptic feedback to the teleoperator. The virtual fixtures in this work are based on Voxmap-PointShell algorithm \cite{hulin2012, Sagardia2018} and more details on their implementation and parameter tuning can be found in \cite{tubiblio109900}.

\section{Experiments and Results}
\label{sec:results}
\subsection{Robust Object Localization: Validation and Analysis}
An object localization approach is taken for 3D visualization and thus, accuracy, run-time and robustness of the proposed algorithm is reported. These results are important as the created virtual reality should closely match the real remote scene. For this, we measure the ground truth of the relative poses between the object and the camera using Vicon tracking system and evaluate the performance on sequences that represent peg-in-hole insertion task (see video attachment). The algorithm is also compared to Apriltag2 \cite{Wang2016} (AP2) and ARToolKitPlus \cite{artoolkit} without (2) and (3) (multiART).
\begin{figure}
\begin{center}
    \includegraphics[width=0.45\textwidth]{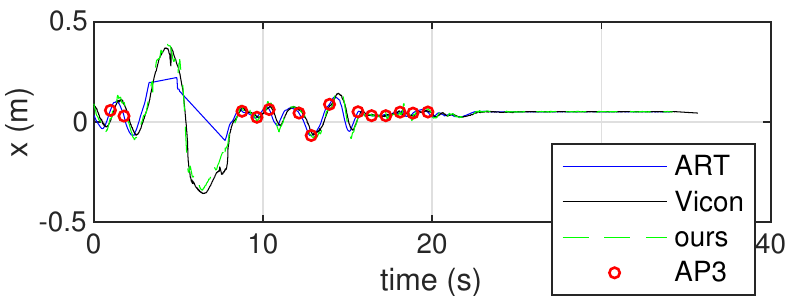}
    \vspace{0.1cm}
    \includegraphics[width=0.45\textwidth]{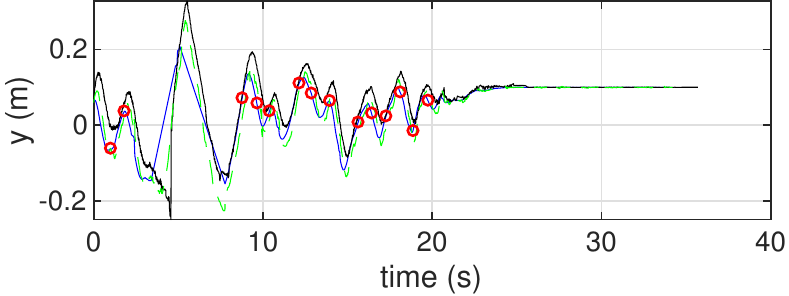}
    \vspace{0.1cm}
    \includegraphics[width=0.43\textwidth]{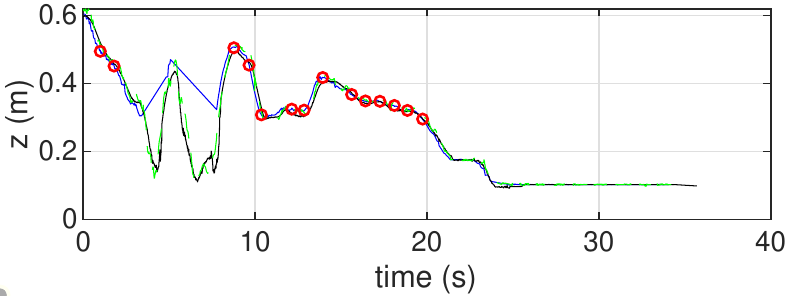}
    \caption{Our proposed algorithm 1 for object tracking (denoted ours) is compared to ground truth (Vicon measurements). The algorithm is compared with two other popular fidicual detection frameworks namely AprilTag 2 (AP2) and ARToolKitPlus (multiART). Our proposed algorithm is robust to losing the fidicuals in an image, and compensates the delay.}
    \label{vicon_results}
  \end{center}
\end{figure}
In Fig. \ref{vicon_results}, estimated trajectories of relative poses are compared with Vicon measurements. As depicted, our proposed algorithm is robust against loss-of-sight problems of object localization with a hand-eye camera while AP2 and multiART produce jumps as no markers are detected (t=3s to t=8s as an example). This is due to the design of the algorithm where we utilize VIO estimates of the camera pose when the marker is not detected. Furthermore, multiART suffers from time delay, while AP2 has both the time delay, and slow run-time. On the other hand, our proposed algorithm compensates the time delay, resulting in accurate estimates. Five experiments have been conducted to determine the accuracy of the selected methods with respect to the ground truth. Note that the trajectory selected includes loss-of-sight and time delay. The corresponding root mean squared errors (RMSE) have been reported in Table \ref{rmse}. However, as seen in Table \ref{rmse}, AP2 is slow while using high-resolution images, and this results in more errors as we compare the trajectories. In our approach, these trajectories are relevant as we aim for creating virtual reality with object localization methods. Within our experiments, the analysis of the accuracy, robustness and run-time further justifies the proposed algorithm and its additional complexity.

\begin{figure*}
  \centering
  \includegraphics[width=1\textwidth]{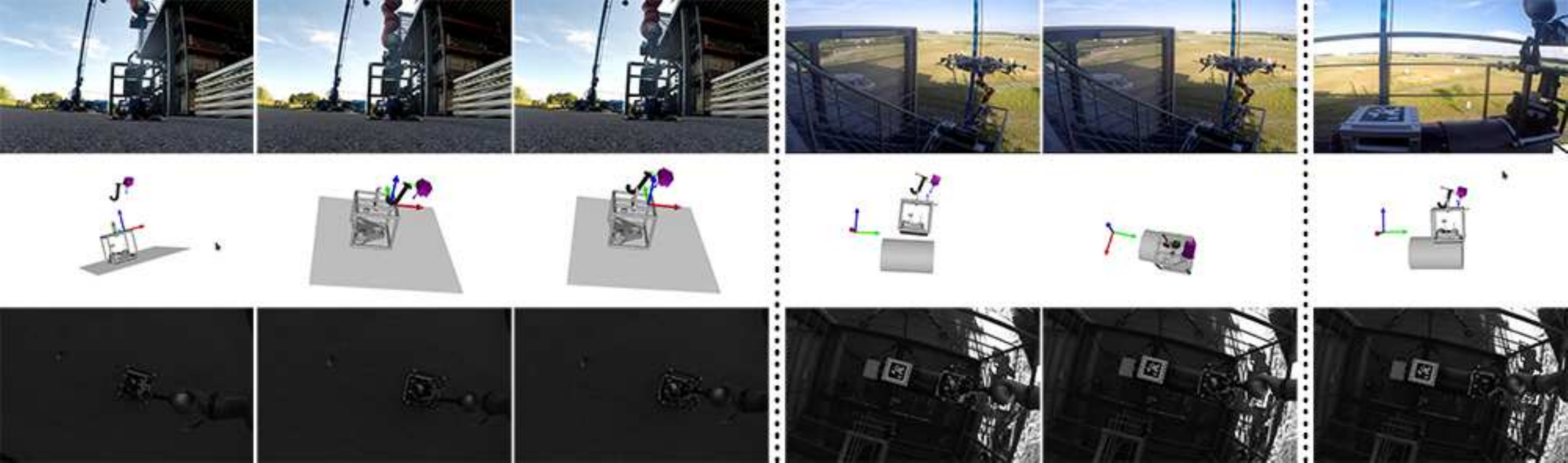}
 \caption{Results of field experiments for AEROARMS \cite{aeroarms} industrial scenario. SAM successfully deployed and retrieved a pipe inspection robot by performing grasping (left), placing (middle) and pressing (right). As we consider outdoor manipulation tasks with an industry relevancy, the system has to both address force feedback, and 3D visual feedback. 2D visual feedback (bottom row), as depicted above, is not sufficient as the depth information is missing and subject to under exposure. On the other hand, the virtual environment (middle row) does not suffer from these problems, and the operator can zoom-in and out, and change its sight-of-view. These experiments show SAM with telepresence as a viable option for future applications.}
  \label{flightexperiments}
\end{figure*}

\begin{table}
\centering
\caption{Accuracy and run-time analysis}
\label{rmse}
\begin{tabular}{|c|c|c|c|} \hline
 & AP2 & multiART+ & ours  \\ \hline
$e_{\text{x,rmse}}$ [m] &  0.1690 & 0.1124  & \textbf{0.0252}  \\ \hline
$e_{\text{y,rmse}}$ [m] & 0.1265  & 0.0847  & \textbf{0.0503}  \\ \hline
$e_{\text{z,rmse}}$ [m] & 0.1308  & 0.077  & \textbf{0.0316}  \\ \hline
$e_{\text{$\phi$,rmse}}$ [rad] & 0.2843  & 0.1867  & \textbf{0.1232}  \\ \hline
$e_{\text{$\theta$,rmse}}$ [rad] & 0.1955 & 0.1232 & \textbf{0.0703}  \\ \hline
$e_{\text{$\psi$,rmse}}$ [rad] & 0.2565 & 0.1755 & \textbf{0.1153}  \\ \hline
$t_{\text{run}}$ [s] & 0.839 $\pm$ 0.0616 & 0.0525 $\pm$ 0.0218 & \textbf{0.0049 $\pm$ 0.013} \\ \hline
\end{tabular}
\end{table}

\subsection{Peg-in-Hole Insertion with Virtual Fixtures}
\begin{figure}
\begin{center}
    \includegraphics[width=0.4\textwidth]{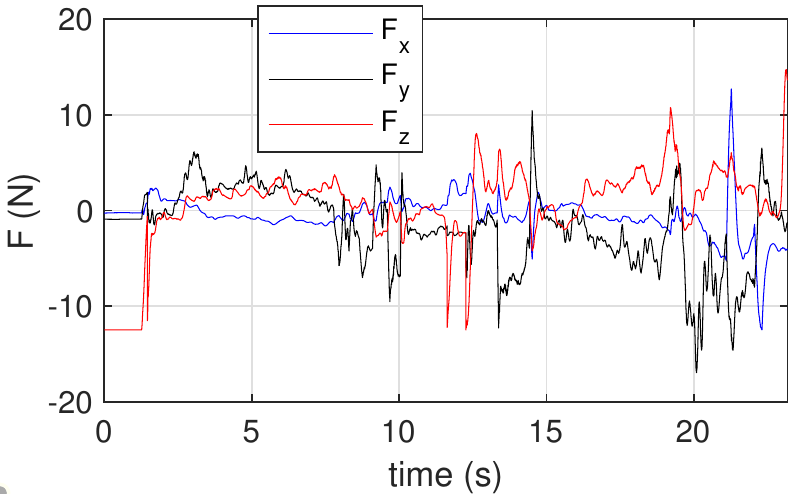}
    \caption{Force signals on slave's end-effector expressed in LWR base frame. These forces compose of artificial force from a virtual fixture, and computed forces from master's commanded positions.}
    \label{peg_in_hole_forces}
  \end{center}
\end{figure}

\begin{figure}
\begin{center}
    \includegraphics[width=0.4\textwidth]{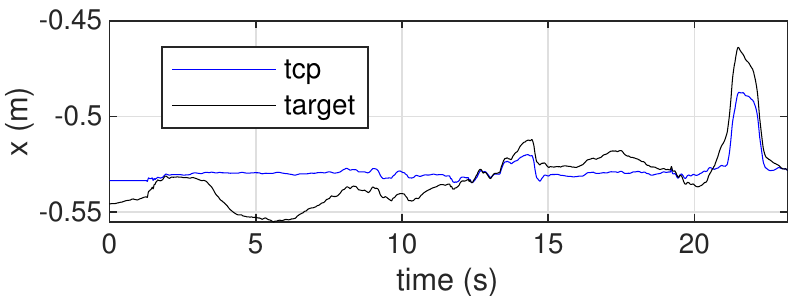}
    \vspace{0.1cm}
    \includegraphics[width=0.4\textwidth]{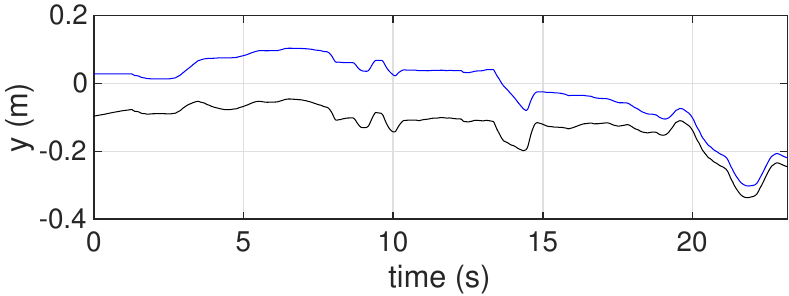}
    \vspace{0.1cm}
    \includegraphics[width=0.4\textwidth]{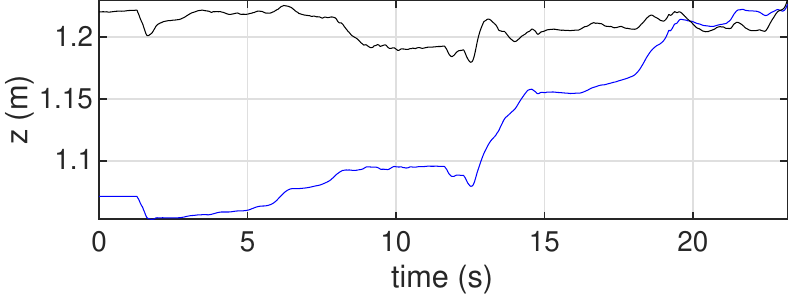}
    \caption{TCP and target positions expressed in LWR base frame. For peg-in-hole insertion, tcp is commanded to follow the target. Note that the target position changes as SAM moves, and it is expressed i in LWR base frame.}
    \label{peg_in_hole_tracking}
  \end{center}
\end{figure}

A peg-in-hole insertion task with margins of error less than 1cm is considered in which operator does not have any direct visual contact to the real scene. The main challenge in this setting is on the fidelity of virtual reality and resulting virtual fixtures. With the fidelity provided by our proposed algorithm and resulting virtual fixtures, a peg-in-hole task has been performed (see the attached video material). The results are depicted in Fig. \ref{peg_in_hole_forces} and Fig. \ref{peg_in_hole_tracking}. Fig. \ref{peg_in_hole_forces} plots force signals acting on the slave end-effector which constitutes computed force from master's position commands, and force due to the virtual fixtures. Position tracking of tcp towards the target (hole) is shown in Fig. \ref{peg_in_hole_tracking}. As these position signals are expressed in LWR base frame (see Fig. \ref{fig:sec3:1} for definition), the target also moves due to the motion of SAM. This experiment shows the benefits of our proposed telepresence system, as SAM is able to perform a precise manipulation task. Note that the accuracy of object localization improves over reported values in Table \ref{rmse} when the peg is near the hole (shown in Fig. \ref{vicon_results}) which makes the task feasible.

\subsection{Field Experiments and Validation}
A field experiment is conducted in order to demonstrate the applicability of SAM within a relevant industrial scenario for aerial manipulation. This scenario involves a maintenance and inspection task in which SAM has to deploy and retrieve a 6.4kg inspection robot to a remotely located pipe. To transport the inspection robot, a cage (approximately of the same size as the pipe and the inspection robot) has been designed. For this mission, SAM has to (a) grasp the cage with a hook at location A with a hook used as end-effector for the LWR, (b) move to location B where the pipe is located, (c) place the cage on the pipe, and (d) press the cage while the inspection robot moves out. The teleoperator is located in a ground station and thus, has no direct visual contact to the scene. For this scenario, we tackle precision grasping, placing and pressing tele-manipulation tasks at a remote location, and the results are depicted in Fig. \ref{flightexperiments}. In particular, 2D images alone do not show the depth information (placing task) and are often occluded (grasping and pressing phases). With only force feedback, a precise manipulation is difficult for this scenario. On the other hand, the virtual reality provides 3D information of the remote scene, and moreover, one can change the sight-of-view to avoid an occluded visual feedback. These results show the benefits of our telepresence system. By touching and seeing, the teleoperator is able to perform precise manipulation tasks for an industrial use-case. 

The field experiments for AEROARMS industrial scenario did not use the haptic guidance using virtual fixtures and VIO compensations for achieving the basic teleoperation tasks. For further improving the inspection and maintenance scenario, we plan to perform a user-study to investigate the degree of improvements with this shared autonomy concept and further joint demonstration with recent developments on SAM \cite{sarkisov20, coelho20}. Lastly, robotic introspection \cite{grimmett2013knowing} for object localization is another research direction that can support in industrial deployments of these systems.

\section{Conclusion}
\label{sec:conclusion}
This paper presents a vision-inertial telepresence concept in which onboard sensors, an object tracking algorithm and databases of objects were utilized to provide a 3D visualization of the scene in real-time. From our experiences in the field, we believe that providing a 3D visual feedback to the tele-operator is required in aerial manipulation applications at remote sites where a direct and close visual contact to the objects are genuinely difficult. Our demonstration of advanced aerial manipulation shows that SAM with telepresence is a viable concept for inspection and maintenance applications.

\section{Acknowledgements}
Special thanks to Michael Vilzmann for the support on FCC, Thomas Hulin for the support on peg-in-hole experiments and Nari Song for the support on video editing.





\bibliographystyle{IEEEtran}
\bibliography{bibliography}

\end{document}